%% file: main.tex
\documentclass[letterpaper,10pt,conference]{ieeeconf}
\IEEEoverridecommandlockouts

\usepackage{amssymb}
\usepackage{booktabs}
\usepackage{color}
\usepackage[backend=biber,
            url=false,
            isbn=false,
            doi=false,
            backref=false,
            style=ieee,
            natbib=true,
            mincitenames=1,
            maxcitenames=1,
            citestyle=numeric-comp,
            sorting=none,
            block=none]{biblatex}
\renewcommand{\bibfont}{\small}
\usepackage{graphicx}
\let\labelindent\relax
\usepackage{enumitem}
\usepackage{amsmath}
\usepackage{wrapfig}
\usepackage{caption}
\usepackage{subcaption}
\usepackage{tabularx,array}
\usepackage{listings}
\usepackage{xcolor}
\usepackage{tcolorbox}
\usepackage{xspace}

\usepackage[colorlinks=true,
            linkcolor=blue,
            citecolor=blue,
            urlcolor=blue]{hyperref}
\usepackage[nameinlink, capitalize]{cleveref} 

\usepackage{siunitx}
\usepackage{caption}
\newcommand{\ACRO}{{DexEXO}\xspace}

\definecolor{lightgray}{gray}{0.95}

\lstdefinelanguage{yaml}{
  morekeywords={true,false,null,yes,no},
  sensitive=false,
  morecomment=[l]{\#},
  morestring=[b]",
  morestring=[b]'
}

\begin{document}
    
\title{\ACRO: A Wearability-First Dexterous Exoskeleton for Operator-Agnostic Demonstration and Learning}

\author{Alvin Zhu$^{1, 2*}$, Mingzhang Zhu$^{3*}$, Beom Jun Kim$^{3}$, Jose Victor S. H. Ramos$^{3}$, Yike Shi$^{1, 2}$, \\ Yufeng Wu$^{3}$, Raayan Dhar$^{1}$, Fuyi Yang$^{1}$, Ruochen Hou$^{3}$, Hanzhang Fang$^{1, 2}$, Quanyou Wang$^{3}$, \\ Yuchen Cui$^{1\dagger}$, and Dennis W. Hong$^{3\dagger}$ \\ \href{https://dexexo-research.github.io/}{dexexo-research.github.io}
\thanks{* denotes equal contribution. $\dagger$ denotes equal advising.}
\thanks{$^{1}$Department of Computer Science, $^{2}$Department of Electrical and Computer Engineering, $^{3}$Department of Mechanical and Aerospace Engineering, UCLA, Los Angeles, CA, USA.}
}

\maketitle

\begin{abstract}
Scaling dexterous robot learning is constrained by the difficulty of collecting high-quality demonstrations across diverse operators. Existing wearable interfaces often trade comfort and cross-user adaptability for kinematic fidelity, while embodiment mismatch between demonstration and deployment requires visual post-processing before policy training. We present \ACRO, a wearability-first hand exoskeleton that aligns visual appearance, contact geometry, and kinematics at the hardware level. \ACRO features a pose-tolerant thumb mechanism and a slider-based finger interface analytically modeled to support hand lengths from 140~mm to 217~mm, reducing operator-specific fitting and enabling scalable cross-operator data collection. A passive hand visually matches the deployed robot, allowing direct policy training from raw wrist-mounted RGB observations. User studies demonstrate improved comfort and usability compared to prior wearable systems. Using visually aligned observations alone, we train diffusion policies that achieve competitive performance while substantially simplifying the end-to-end pipeline. These results show that prioritizing wearability and hardware-level embodiment alignment reduces both human and algorithmic bottlenecks without sacrificing task performance. 
\end{abstract}

\input{sections/S0_Introduction}
\input{sections/S1_Related_work}
\input{sections/S2_Hardware_Design}

\input{sections/S3_Data_Collection}

\input{sections/S4_Results}
\input{sections/Limitations}
\input{sections/S6_Conclusion}

\renewcommand*{\bibfont}{\footnotesize}
\printbibliography

\end{document}

%% file: sections/S0_Introduction.tex
\section{Introduction}
\label{sec:intro}

Learning robust dexterous manipulation remains fundamentally limited by the availability of scalable, high-fidelity demonstrations that capture the closed-loop, contact-rich strategies humans employ in daily tasks \cite{dapg,openai_cube,vima,handeye,aloha}. Although recent advances in robot learning have shown strong gains from larger and more diverse human datasets, collecting such data for multi-finger hands remains particularly difficult due to high-dimensional kinematics, frequent occlusions, and complex hand–object contact dynamics~\cite{robohive,ye2026scirobotics_vtpretrain,liu2024realdex,huang2025review_humanoid_dexterity,tanaka2025scaler,welte2025interactive_imitation,open_x_embodiment_rt_x_2023,rt22023arxiv}. In contrast to parallel-jaw grippers, where portable teaching interfaces scale effectively~\cite{chi2024umi_rss}, high-DoF hands continue to rely on interfaces that trade off naturalness, wearability, and motion fidelity, especially in the thumb, whose abduction, adduction, and opposition enable complex in-hand manipulation.

\begin{figure}[htbp]
\centering
\includegraphics[width=0.99\linewidth]
{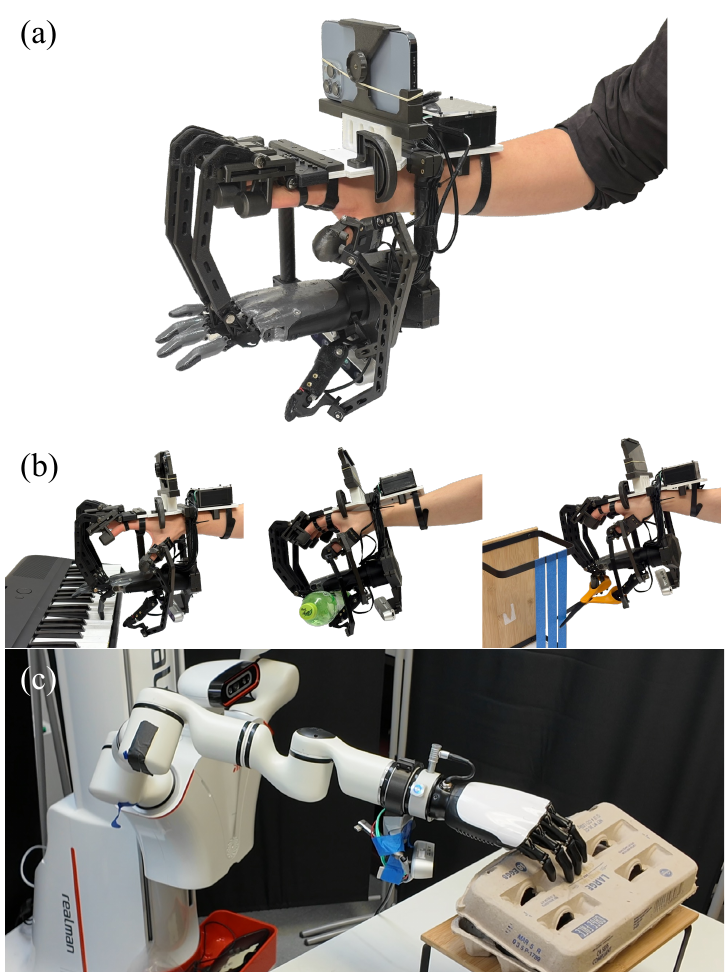}
\captionsetup{belowskip=-10pt}
\caption{\textbf{System overview of DexEXO.} (a) Full device worn on a user's hand. (b) Demonstrations of piano playing, full-hand grasping, and scissors cutting. (c) Policy deployment on the robot.}
\label{fig:Teaser}
\end{figure}

Existing sources of dexterous demonstrations generally fall into three categories: (i) \textbf{simulation and videos}, (ii) \textbf{robot teleoperation}, and (iii) \textbf{wearable interfaces} such as gloves and exoskeletons. Video and simulation data scale efficiently and broadly~\cite{shaw2024_internet_handmotion,singh2024_hoi_pretrain,aura,taccel2025, r2r2r, wang2024dexcap,realtosimtoreal}, yet accurately capturing contact forces, fine hand–object interactions, and transferring them to hardware remains challenging~\cite{taccel2025,sim2real_force_2026, zhao2025polytouchrobustmultimodaltactile}. Teleoperation provides demonstrations directly in the robot control space~\cite{handa2020dexpilot_arxiv,he2024omnih2o,fu2024_mobilealoha,10801581}, but dexterous hand teleoperation is often slow, unintuitive, costly to scale, and limited by insufficient haptic feedback for contact-rich manipulation~\cite{handa2020dexpilot_arxiv}. Wearable devices improve embodiment by mechanically coupling human motion to the robot, reducing retargeting ambiguity and enabling more natural demonstrations~\cite{xu2025dexumi,fang2025dexop_arxiv,du2025mile_arxiv}. However, prior work shows that wearables can introduce a visual embodiment gap during data collection, requiring additional post-processing before training~\cite{xu2025dexumi}. These systems also frequently sacrifice comfort for fidelity and can be difficult to fit across users due to anthropometric variation~\cite{ge2023_selfalign_exo,brogi2024_thumb_module}.

Motivated by these limitations, we present \ACRO, a wearable hand exoskeleton for data collection designed around two principles: \textbf{wearability and cross-user adaptability} to enable scalable, sustained demonstration collection, and \textbf{aligned visual and kinematic embodiment} for an efficient end-to-end pipeline from demonstration to policy training. Our approach preserves natural thumb and finger behaviors during data collection while maintaining a consistent, learnable mapping to the target robot hand. By incorporating a passive hand, the wrist-mounted camera view aligns with that of the physical robotic hand, eliminating visual discrepancies between collection and inference. The design targets practical deployment scenarios in which operators perform repeated tasks across diverse environments with minimal setup, while retaining the motion and visual fidelity required for learning manipulation skills~\cite{fang2025dexop_arxiv,du2025mile_arxiv,si2024telehand_arxiv}. In summary, our contributions are:

\begin{itemize}\setlength\itemsep{0.15em}
\item A wearability-first hand exoskeleton with analytically validated anthropometric compatibility, enabling cross-user operation without rigid alignment or calibration.
\item A pose-tolerant thumb mechanism that preserves the natural human thumb workspace while maintaining a consistent, controllable mapping to robot thumb DOFs.
\item An embodiment-aligned data collection and policy training pipeline that eliminates segmentation and visual post-processing, enabling direct learning from only raw wrist-mounted RGB observations.
\end{itemize}

\begin{figure*}[t!]
 \centering
 \includegraphics[width=0.9\linewidth]{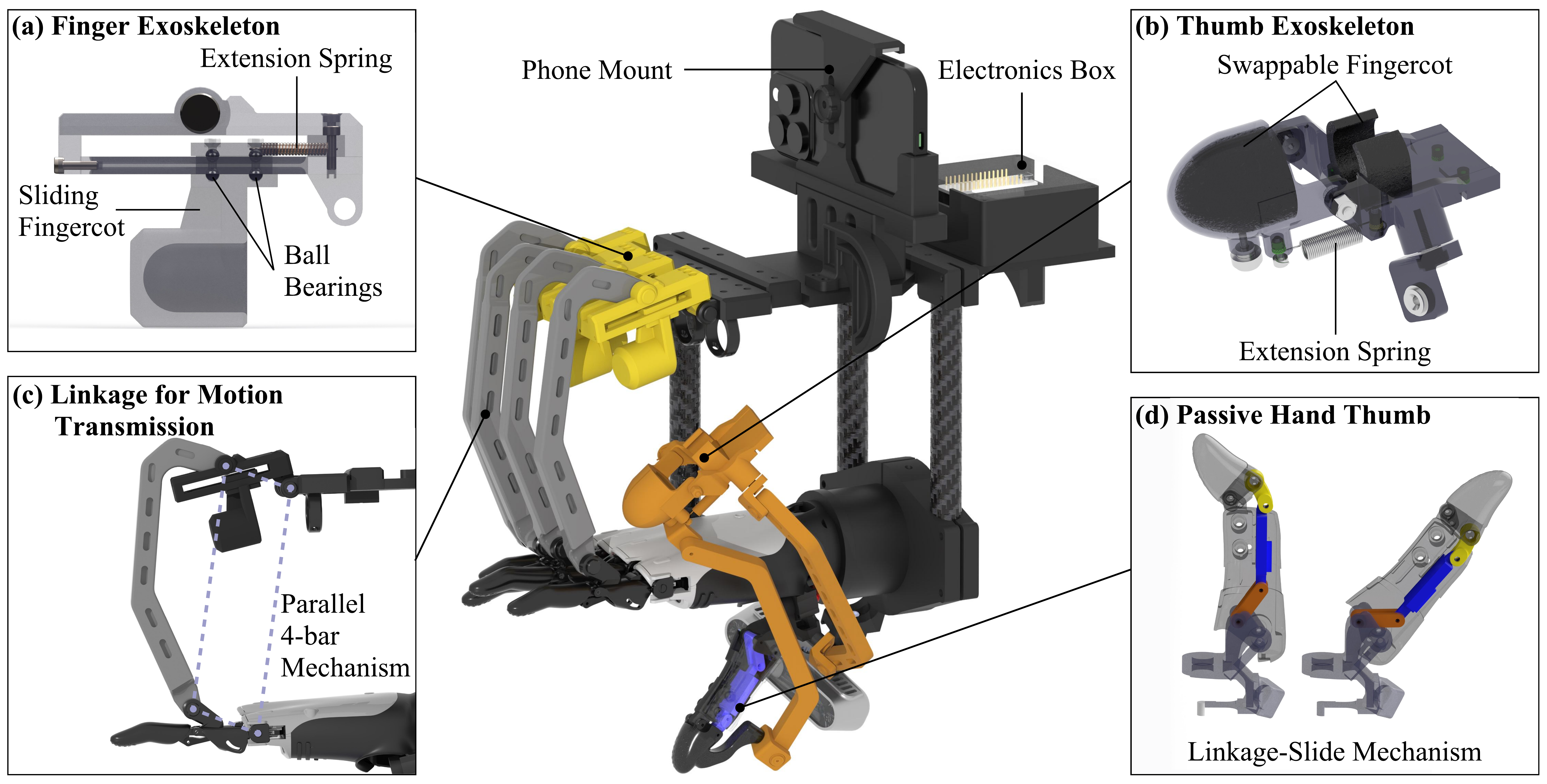}
 \captionsetup{belowskip=-10pt}
 \caption{Mechanical overview of \ACRO. \ACRO integrates a linkage-driven wearable exoskeleton, a passive data-capture hand, and an onboard sensing/power module. Insets highlight key subsystems: (a) passive finger slider for cross-user fit, (b) pose-tolerant thumb coupling interface, (c) parallel four-bar finger linkage for motion transmission, and (d) passive hand thumb that reproduces the intended thumb DOF}
 \label{fig:me_big_figure}
\end{figure*}

%% file: sections/S1_Related_work.tex
\section{Related Work}
\label{sec:related}

\subsection{Teleoperation for Dexterous Manipulation}

Teleoperation remains the dominant approach for collecting high-quality dexterous demonstrations, but existing interfaces involve inherent trade-offs. Vision-based systems provide an unencumbered user experience, yet they are fundamentally limited by line-of-sight occlusion and tracking instability during contact-rich interactions~\cite{handa2020dexpilot_arxiv,shaw2024_internet_handmotion, bytedance2025bytedexter}. Data gloves alleviate these tracking challenges and can provide tactile feedback~\cite{ruppel2024_tactile_glove}, but they introduce the "correspondence problem"~\cite{retarget2025_unified}. Without physical constraints enforcing robot kinematics on the human hand, glove-based demonstrations may generate trajectories that are kinematically infeasible for the target robot, despite recent advances in retargeting algorithms~\cite{mandi2025_dexmachina_arxiv,retarget2025_ultrafast}.

\subsection{Learning from Human Hand Videos}

To circumvent the hardware limitations of teleoperation, recent work has explored learning directly from large-scale human video data~\cite{shaw2024_internet_handmotion,singh2024_hoi_pretrain,dexman2025_openreview, Liu_2022_CVPR, chao_cvpr2021}. These approaches exploit the scale of internet videos to acquire rich visual and geometric priors for hand--object interaction~\cite{shaw2023videodex,mandikal2022dexvip, bharadhwaj2024track2act,grauman2022ego4dworld3000hours}. However, video-based learning faces a "physicality gap," as it lacks explicit information about contact forces and closed-loop interaction dynamics~\cite{sim2real_force_2026,taccel2025}. As a result, policies trained purely from video often struggle to transfer directly to physical systems and typically require additional fine-tuning on contact-rich demonstrations via teleoperation~\cite{ye2026scirobotics_vtpretrain,liu2025_vtdexmanip_iclr}.

\subsection{Exoskeleton and Mechanically Coupled Interfaces}

Exoskeleton and mechanically coupled interfaces aim to reduce embodiment mismatch by physically linking human motion to robot kinematics, improving controllability compared to loosely coupled systems~\cite{du2025mile_arxiv,si2024telehand_arxiv, zhang2025doglove}. Recent systems such as \textbf{DexUMI}~\cite{xu2025dexumi} pursue scalable in-the-wild data collection through a lightweight wearable and vision-based reconstruction. However, their rigid exoskeleton geometry, derived from non-anthropomorphic robotic hand proportions, provides limited adaptation to diverse human hand sizes, increasing joint-alignment sensitivity across operators and potentially constraining ergonomic tolerance during sustained use~\cite{zhang2026human}. Additionally, visual embodiment mismatch requires segmentation and inpainting prior to policy training. 
Devices such as \textbf{DexOP}~\cite{fang2025dexop_arxiv} address embodiment alignment through hardware--robot co-design, employing linkage-driven passive mechanisms to enforce strong kinematic correspondence between the operator and robot hand. While this tight coupling improves demonstration fidelity, it binds the interface to specific robot geometries, limiting adaptability to diverse hand designs already used at scale. Moreover, rigid linkage constraints reduce tolerance to anthropometric variation and restrict residual thumb motion, constraining the natural abduction, adduction, and opposition workspace required for complex in-hand manipulation~\cite{ge2023_selfalign_exo,brogi2024_thumb_module,wang2025_icor_exo}. 
In contrast, our approach targets the intersection of wearability-first extended use and robust mapping by incorporating a pose-tolerant thumb mechanism that preserves natural thumb motion without sacrificing controllability.

%% file: sections/S2_Hardware_Design.tex
\section{Hardware Design}
\label{sec:hardware}
\subsection{Hardware Overview}
\ACRO comprises (i) a linkage-driven wearable exoskeleton, (ii) a passive demonstration hand, and (iii) an onboard sensing and power module for untethered operation. The passive hand follows the geometry of the 6-DoF OYMotion ROH-AP001 (ROHand) \cite{oymotion_roh_ap001}, featuring a 2-DoF thumb (IP flexion/extension and TM abduction/adduction) and single-DoF flexion for each finger. As shown in Fig.~\ref{fig:me_big_figure}, the exoskeleton transmits operator motion through two coupling architectures. The four fingers use parallel linkage mechanisms that provide identical flexion/extension correspondence while accommodating inter-user variation. The thumb employs a multi-DoF coupling that allows the exoskeleton structure to translate and rotate relative to the palm while still transferring the key thumb motions to the passive hand, improving comfort and adaptability across hand sizes. A dorsal-mounted electronics module supplies onboard power and data logging, while a wrist-mounted iPhone provides pose capture for in-the-wild data collection.

\begin{figure}[t!]
 \centering
 \includegraphics[width=0.8\linewidth]{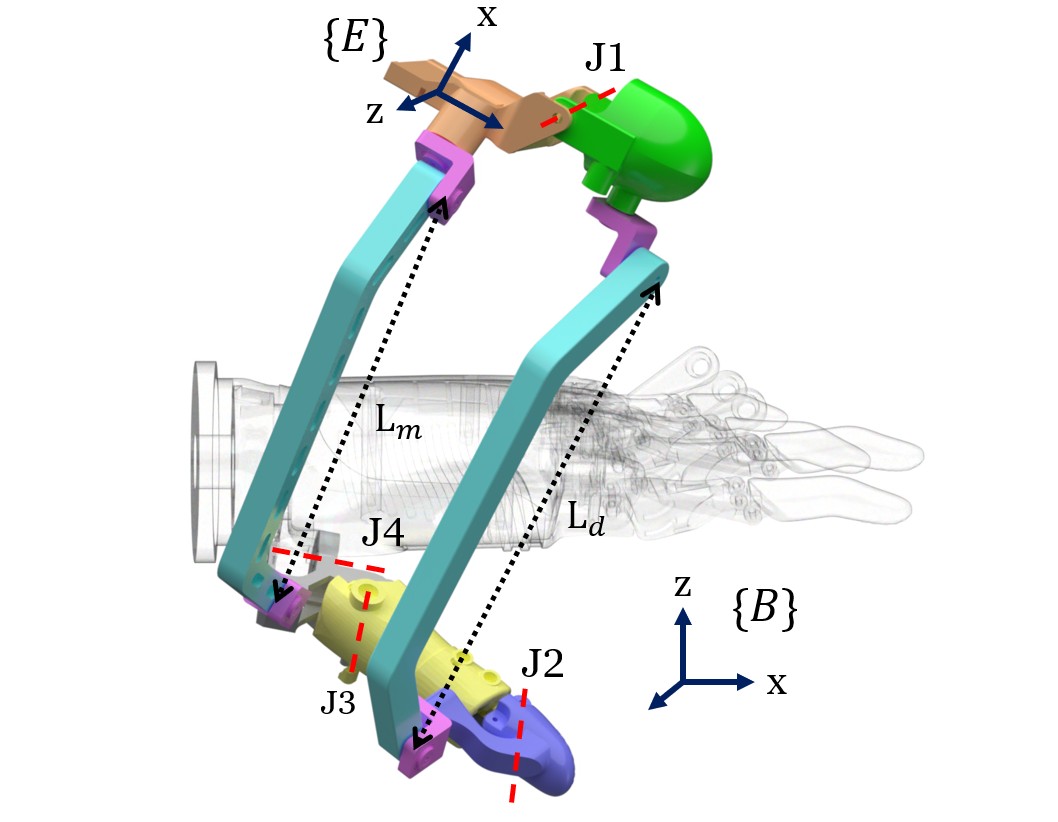}
 \captionsetup{belowskip=-10pt}
 \caption{Kinematic schematic of the exoskeleton thumb and its distal and metacarpal linkages. The four swivel joints (shown in purple) allow self-alignment between the exoskeleton frame ${E}$ and the palm base frame ${B}$ while maintaining fixed linkage lengths $L_d$ and $L_m$.}
 \label{fig:thumb_mech}
\end{figure}

\subsection{Passive Hand}
 To ensure high-fidelity proprioception, \ACRO integrates six joint encoders within a rigid, rib-reinforced mounting structure that prevents misalignment, backlash, and sensor drift, maintaining stable joint-angle measurements during dynamic manipulation. In parallel, we performed kinematic identification using a URDF to design a custom linkage-slide mechanism that matches the actual hand kinematics, ensuring consistent and physically accurate trajectory mapping.

\subsection{Wearability-First Design for Cross-Operator Deployment}
\label{sec:wearability}

\ACRO is designed for deployment without rigid joint alignment or per-user calibration. Instead of enforcing strict geometric coincidence between human and exoskeleton joints, passive tolerance mechanisms absorb anthropometric variation locally while preserving structured motion transfer.

\subsubsection{Slider-Based Finger Interface}
Each finger employs a passive spring-loaded linear slider coupled to a compliant fingercot. The slider fits variation in finger length while preserving sufficient curl displacement, decoupling insertion depth from joint-axis alignment and improving fit robustness.

\begin{figure*}[t!]
 \centering
 \includegraphics[width=0.93\linewidth]{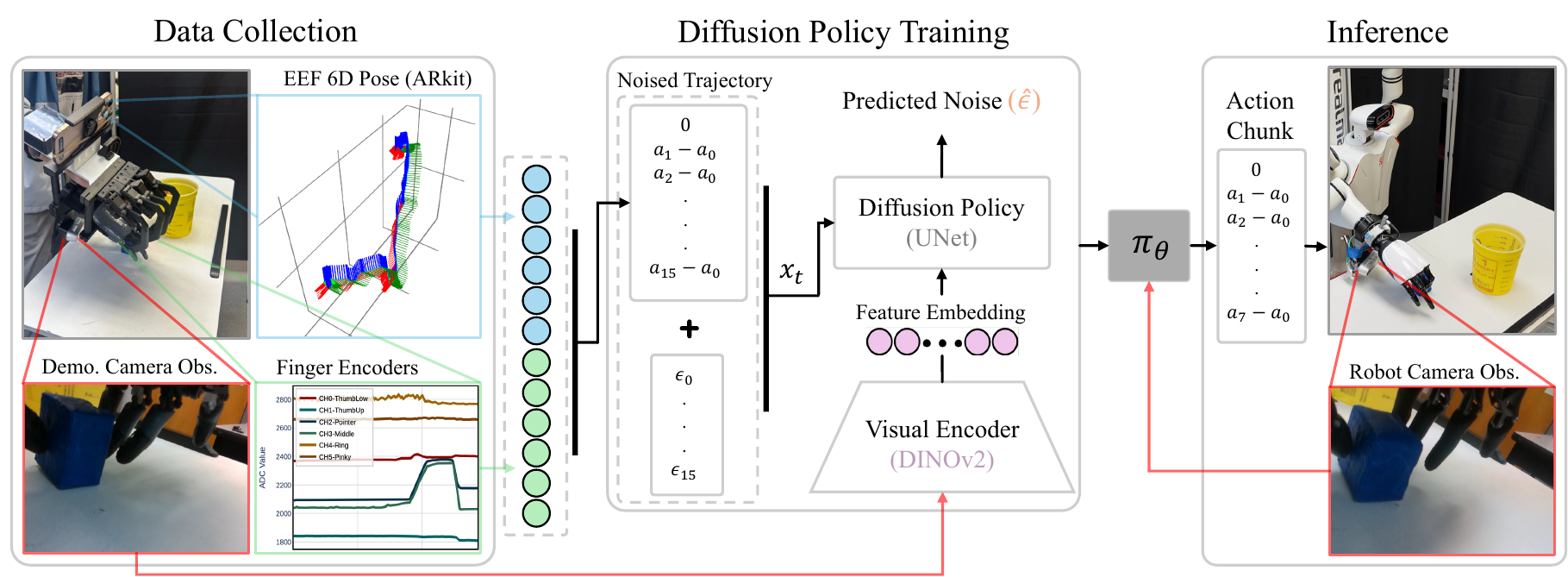}
 \captionsetup{belowskip=-10pt}
 \caption{An overview of the full demonstration data modalities, policy training, and inference with visual-aligned observations.}
 \label{fig:overview}
\end{figure*}

To estimate the range of compatible hand sizes, we analyze the mechanical limits of the slider using the middle finger as the governing digit. Let $L_{min}$ denote the minimum TPU ring--to--fingercot distance at rest, $d_{max}$ the maximum distance permitted by slider travel, $d_{curl}$ the minimum free slider length required for full finger flexion, $\delta$ the maximum allowable offset of the TPU ring above the webbing, and $r$ the middle-finger-to-hand-length ratio ($r \approx 0.39$--$0.40$) reported in~\cite{buryanov2010hand}. The compatible bounds are: 

\begin{equation}
L_{max} = d_{max} - d_{curl}, \qquad
MFL_{max} = L_{max} + \delta
\label{eq:slider_bounds}
\end{equation}

\begin{equation}
H_{min} = \frac{L_{min}}{r}, \qquad
H_{max} = \frac{MFL_{max}}{r}
\label{eq:hand_bounds}
\end{equation}

Placement variability is incorporated by allowing the TPU ring to sit up to $\delta = 17$\,mm above the webbing before restricting PIP flexion. Even accounting for the ring width (8\,mm), sufficient clearance remains for natural finger curl.

Substituting $L_{min} = 56$\,mm, $d_{max} = 86$\,mm (30\,mm travel), $d_{curl} = 16$\,mm, $\delta = 17$\,mm, and $r = 0.40$ yields $L_{max} = 70$\,mm, $MFL_{max} = 87$\,mm, and a compatible hand-length range of $H_{min} = 140$\,mm to $H_{max} = 217$\,mm. All participants in our user study ($n = 14$, 165--195\,mm) fall within this range, validating cross-user compatibility.

\subsubsection{Swappable Compliant Thumb Interface}
Complementing the pose-tolerant linkage, the thumb employs a swappable TPU fingercot coupled to the distal linkage through compliant elements. Unlike rigid shells requiring precise axis alignment, the soft fingercot accommodates variation in thumb length and joint-center location while maintaining stable contact during pinch and grasp.

\subsection{Pose-tolerant Thumb Mechanism}
\label{sec:thumb}
The thumb is challenging for wearable interfaces due to inter-user anatomical variability, where rigid axis alignment can cause discomfort and restrict motion. \ACRO addresses this with a pose-tolerant thumb coupling that preserves wearability while remaining functional with the robotic hand’s IP flexion/extension and TM ab/ad configuration.

\subsubsection{Mechanism overview}
As illustrated in Fig.~\ref{fig:thumb_mech}, the exoskeleton thumb contains an instrumented IP joint $J_1$ with angle $\theta_1$. The passive thumb includes the IP joint $J_2$ with angle $\theta_2$ and the TM ab/ad joint $J_4$ with angle $\theta_4$, where $J_3$ is mechanically coupled to $J_2$. The exoskeleton thumb is connected to the passive thumb through two rigid linkages: a distal linkage and a metacarpal linkage. This architecture avoids enforcing rigid orientation alignment between the exoskeleton and the human thumb. Instead, only geometric distance constraints are imposed, enabling the exoskeleton to translate and rotate relative to the palm while remaining mechanically coupled.

\subsubsection{Simplified kinematic model}
Let $\{B\}$ and $\{E\}$ denote a palm-base frame and an exoskeleton frame. The relative pose of the exoskeleton with respect to the palm is

\begin{equation}
{}^{B}T_{E} =
\begin{bmatrix}
{}^{B}R_{E} & {}^{B}p_{E} \\
0 & 1
\end{bmatrix}
\in SE(3),
\label{eq:thumb_pose}
\end{equation}.

Denote the passive thumb configuration as
\begin{equation}
q_p \triangleq [\theta_2~~\theta_4]^\top,
\qquad \theta_3 = f(\theta_2),
\label{eq:thumb_qp}
\end{equation}
and let ${}^{B}r_d(q_p)$ and ${}^{B}r_m(q_p)$ be the distal and metacarpal attachment points expressed in $\{B\}$, computed from the passive hand kinematics.

The corresponding attachment points on the exoskeleton are constant vectors ${}^{E}\bar r_d$ and ${}^{E}\bar r_m$ expressed in $\{E\}$. Their positions in $\{B\}$ are

\begin{equation}
{}^{B}r^{E}_{i} =
{}^{B}R_E \, {}^{E}\bar r_i + {}^{B}p_E,
\qquad i \in \{d,m\}.
\label{eq:thumb_points}
\end{equation}

The two-link coupling imposes holonomic distance constraints

\begin{equation}
\| {}^{B}r^{E}_{d} - {}^{B}r_{d}(q_p) \| = L_d,
\qquad
\| {}^{B}r^{E}_{m} - {}^{B}r_{m}(q_p) \| = L_m,
\label{eq:thumb_constraints}
\end{equation}

\noindent where $L_d$ and $L_m$ are the distal and metacarpal link lengths.

\subsubsection{Residual pose freedom and self-alignment.}
The exoskeleton pose ${}^{B}T_E$ has six degrees of freedom, while
Eqs.~\eqref{eq:thumb_constraints} impose two independent scalar holonomic constraints under typical thumb configurations. Consequently, for a fixed passive thumb posture $q_p$, the coupled system generically admits a four-dimensional self-motion manifold in the exoskeleton pose space. This residual freedom corresponds to the experimentally observed “wiggle space,” in which the exoskeleton body can translate and rotate relative to the palm without altering the passive thumb posture.

%% file: sections/S3_Data_Collection.tex
\section{Data Collection and Policy Training}
\label{sec:data collection}

\subsection{Data Collection}

\subsubsection{Finger Position Data}

Finger joint positions are measured using six analog encoders embedded within the exoskeleton mechanism. Encoder values are sampled by an onboard microcontroller at 1 kHz and streamed to a host computer using a lightweight binary protocol. We account for the target hand's non-linear actuation kinematics by mapping the exoskeleton encoder data to actuator commands using piecewise linear interpolation across waypoints sampled at identical physical postures on both \ACRO and ROHand.

\subsubsection{End-Effector Pose}

The 6-DOF end-effector pose is captured using an iPhone-based AR tracking system through the TeleDex application \cite{rayyan2026teledex}. Pose data, consisting of position and orientation, is streamed to the host computer and resampled to a fixed 60 Hz rate to ensure consistent timing.

\subsubsection{Visual Observations}

A wrist-mounted Intel RealSense camera records RGB images at $640 \times 480$ resolution and 30 Hz. Each frame is timestamped and stored for downstream policy training.

\subsubsection{Time Synchronization}

All sensor modalities are temporally aligned using video timestamps as the master reference. Asynchronous encoder and pose measurements are matched via nearest-neighbor association.

\begin{table*}[t]
    \centering
    \caption{\textbf{Summary of quantitative results in user study (mean $\pm$ SEM).} Bold indicates best performance per metric.}
    \renewcommand{\arraystretch}{1.2}
    \footnotesize
    \setlength{\tabcolsep}{6.8pt}
    \begin{tabular}{l cc cc cc cc}
    \toprule
    & \multicolumn{2}{c}{\textbf{Scissors Cutting}} 
    & \multicolumn{2}{c}{\textbf{Page Flipping}} 
    & \multicolumn{2}{c}{\textbf{Cup Stacking}} 
    & \multicolumn{2}{c}{\textbf{Piano Playing}} \\
    \cmidrule(lr){2-3} \cmidrule(lr){4-5} \cmidrule(lr){6-7} \cmidrule(lr){8-9}
    \textbf{Method}
    & Success Rate
    & Time (s)$^{\dagger}$ 
    & Success Rate
    & Time (s)$^{\dagger}$
    & Success Rate
    & Time (s)$^{\dagger}$
    & Success Rate
    & Time (s)$^{\dagger}$ \\
    \midrule
    \ACRO 
    & $\mathbf{0.79 \pm 0.10}$
    & $\mathbf{11.7 \pm 1.4}$
    & $\mathbf{0.88 \pm 0.03}$
    & $5.4 \pm 0.6$
    & $\mathbf{0.82 \pm 0.07}$
    & $12.0 \pm 1.1$
    & $\mathbf{0.96 \pm 0.02}$
    & $\mathbf{21.6 \pm 1.8}$ \\
    
    DexUMI
    & $0.00 \pm 0.00$
    & ---
    & $0.86 \pm 0.04$
    & $\mathbf{4.7 \pm 0.7}$
    & $0.80 \pm 0.07$
    & $\mathbf{8.9 \pm 1.0}$
    & $0.62 \pm 0.13$
    & $25.9 \pm 2.5$ \\
    
    Teleoperation
    & $0.00 \pm 0.00$
    & ---
    & $0.51 \pm 0.06$
    & $18.0 \pm 2.1$
    & $0.33 \pm 0.09$
    & $68.6 \pm 13.1$
    & $0.60 \pm 0.09$
    & $97.4 \pm 7.8$ \\
    
    \bottomrule
    \end{tabular}
    \begin{minipage}{\linewidth}
    \vspace{0.3em}
    \footnotesize
    \raggedright
    $^{\dagger}$ Completion time was defined as the average time from picking up scissors 
    to finishing the cut, time for 5 page flips, average time for a successful 
    3-cup stack, and time to play 16 piano notes, respectively.\\
    \end{minipage}
    
    \label{tab:quant_results}
\end{table*}

\begin{figure}[t]
 \centering
 \includegraphics[width=0.99\linewidth]{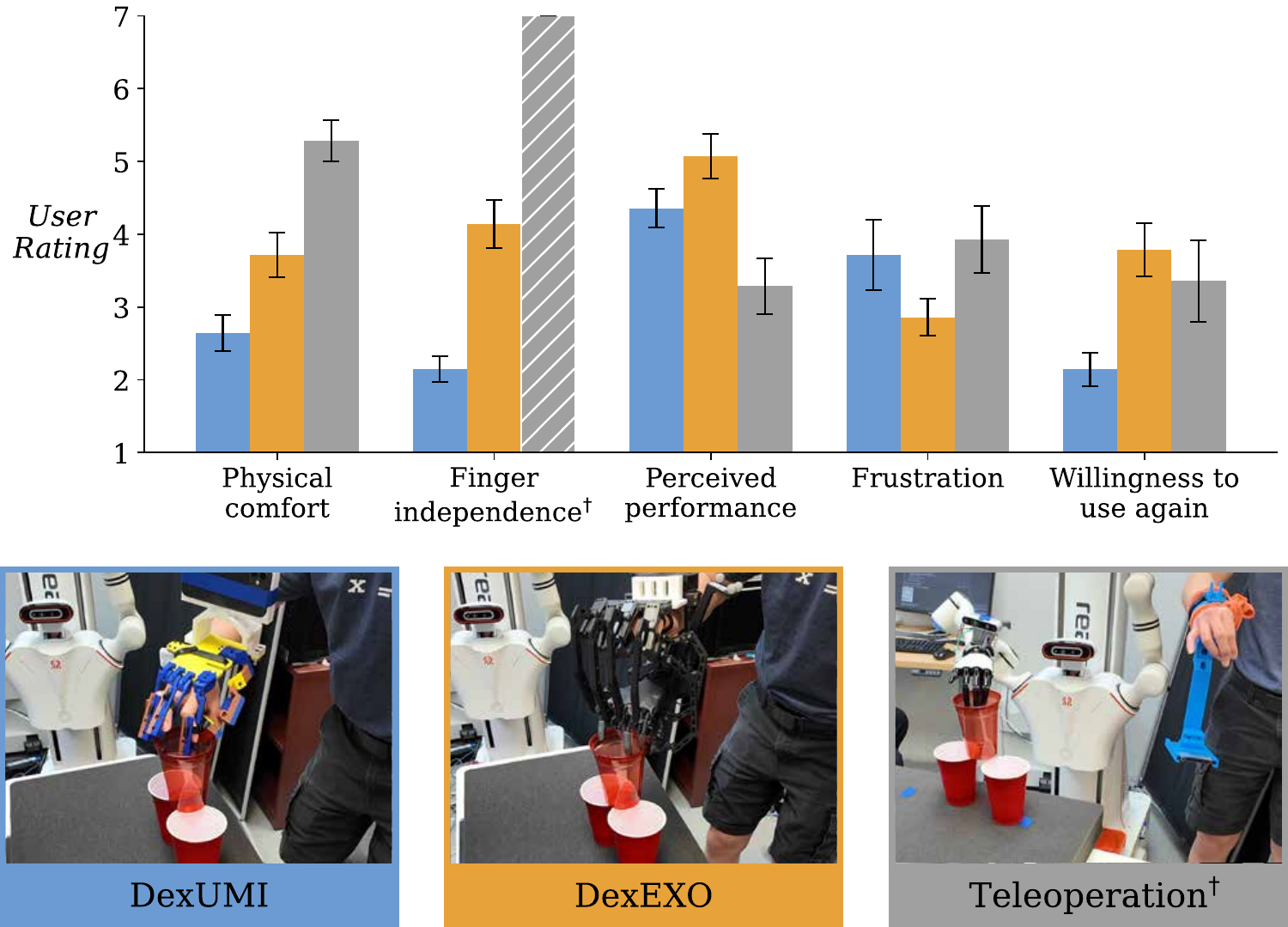}
 \captionsetup{belowskip=-5pt}
 \caption{Subjective feedback results in user study (mean $\pm$ std.). $^{\dagger}$ Finger independence is not applicable to teleoperation, as the user’s natural hand motion is unconstrained.}
 \label{fig:subjective_rating}
\end{figure}

\subsection{Policy Architecture and Training Setup}

Our aligned embodiment enables an efficient pipeline from demonstration to policy training. In particular, the passive hand ensures that the wrist-mounted camera observes a hand geometry consistent with the deployed robotic hand, eliminating the visual embodiment gap that typically necessitates segmentation, masking, or inpainting \cite{xu2025dexumi}. As a result, policies are trained directly from raw wrist RGB observations paired with synchronized end-effector and finger signals.

\paragraph{Observations}
Each training sample includes an RGB frame from the wrist-mounted camera and (optionally) a low-dimensional hand state. The RGB image is resized to $240 \times 240$, randomly cropped to $224 \times 224$, and augmented with color jitter during training. Visual features are extracted using a DINOv2 ViT-S/14 encoder \cite{dinov2}, and the resulting embedding is used as the primary conditioning signal for the policy. When used, the hand state is the 6D absolute finger pose.

\paragraph{Actions}
The policy outputs a 12D action consisting of a 6-DoF end-effector command and 6 finger commands. We train a diffusion policy \cite{diffpolicy} to predict an action horizon of 16 steps and execute the first 8 actions in a receding-horizon manner at inference time. Actions are expressed relative to the initial state of the horizon: the $k$-th predicted action corresponds to $T_k - T_0$. This representation supports reactive closed-loop control while retaining multi-step prediction capability.

All policies in this work use the same diffusion policy backbone and vision encoder; differences between action parameterizations and conditioning signals are evaluated in Sec.~\ref{sec:Results}.

%% file: sections/S4_Results.tex
\section{Results}
\label{sec:Results}


\begin{figure}[t]
 \centering
 \includegraphics[width=0.99\linewidth]{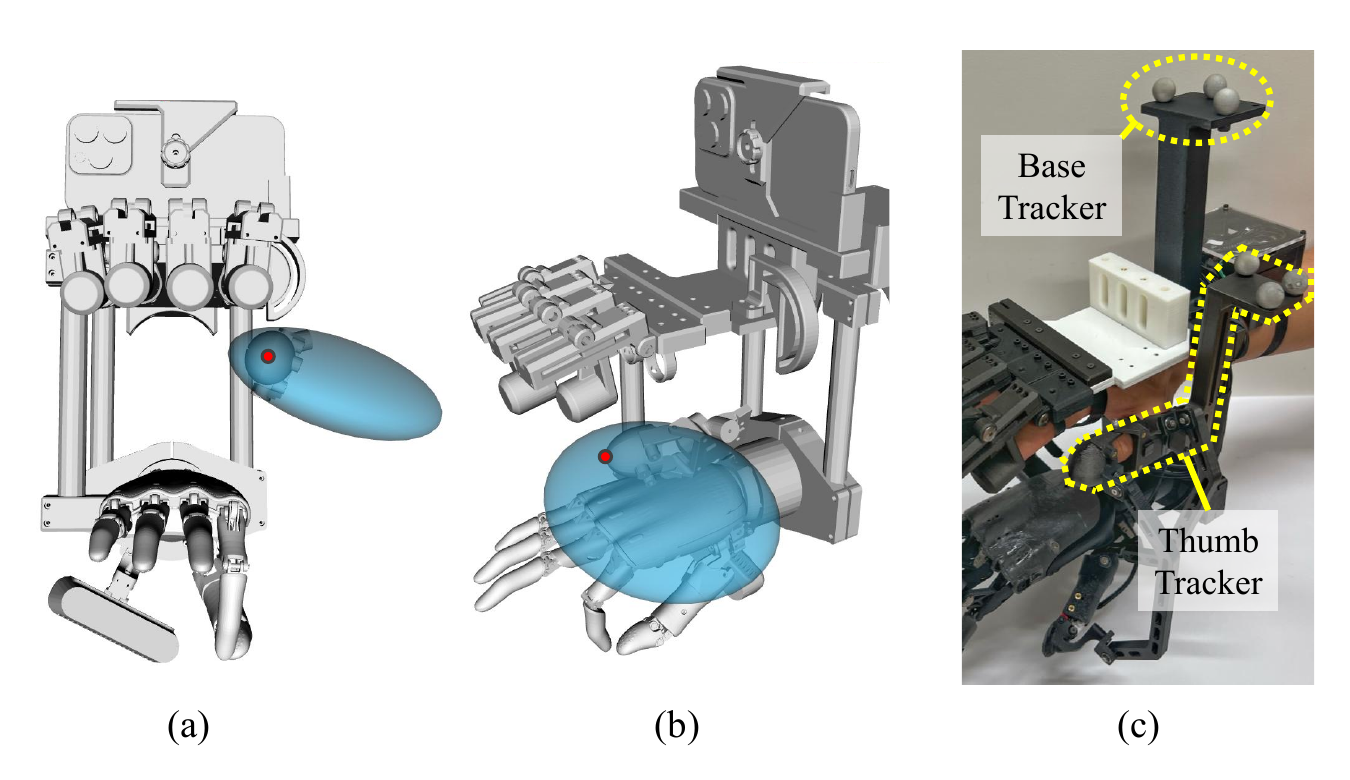}
 \captionsetup{belowskip=-5pt}
\caption{Wiggle-space evaluation of the thumb interface. 
(a) and (b) Two views of the measured wiggle space. The fitted ellipsoid represents the covariance envelope of the sampled trajectory. The red point indicates the nominal fingertip position within the ellipsoid. 
(c) Experimental setup with reflective markers attached for motion-capture measurement.}
 \label{fig:wiggle_space}
\end{figure}

\subsection{Experimental Validation of Thumb Wiggle Space}
The residual pose freedom predicted in Sec.~\ref{sec:thumb} manifests as a self-motion manifold of the exoskeleton relative to the base. As shown in Fig.~\ref{fig:wiggle_space} (c), to experimentally characterize the extent of this allowable motion, we conducted a wiggle-space experiment with the hand maintained in a pinch configuration. Reflective markers were attached to the base and the exoskeleton thumb linkage, and their relative pose was recorded using a motion-capture system for approximately 25\,s while the user performed small natural adjustments within the interface.

The sampled marker positions were expressed in the base frame to obtain the relative motion between the hand and the exoskeleton. The resulting point cloud represents the allowable configuration space (“wiggle space”) during pinch interaction. To summarize the spatial distribution of this motion, we fitted a 3-D ellipsoid to the sampled points using the covariance of the trajectory:

\[
\Sigma = \frac{1}{N-1}\sum_{i=1}^{N}(p_i-\bar{p})(p_i-\bar{p})^\top,
\]
where $p_i \in \mathbb{R}^3$ are the measured positions and $\bar{p}$ is the mean. The ellipsoid axes are obtained from the eigenvalues $\lambda_i$ of $\Sigma$ as
\[
a_i = k\sqrt{\lambda_i},
\]
where $k=2$ corresponds to approximately 95\% coverage.

The fitted ellipsoid has semi-axis lengths of $66.12\,\mathrm{mm}$, $49.19\,\mathrm{mm}$, and $21.14\,\mathrm{mm}$. The large ellipsoidal volume indicates that the interface tolerates substantial variation in thumb placement while maintaining stable kinematic coupling. This tolerance enables the mechanism to accommodate inter-user variation in thumb morphology and placement without requiring precise anatomical alignment.

\begin{figure}[t!]
 \centering
 \includegraphics[width=0.99\linewidth]{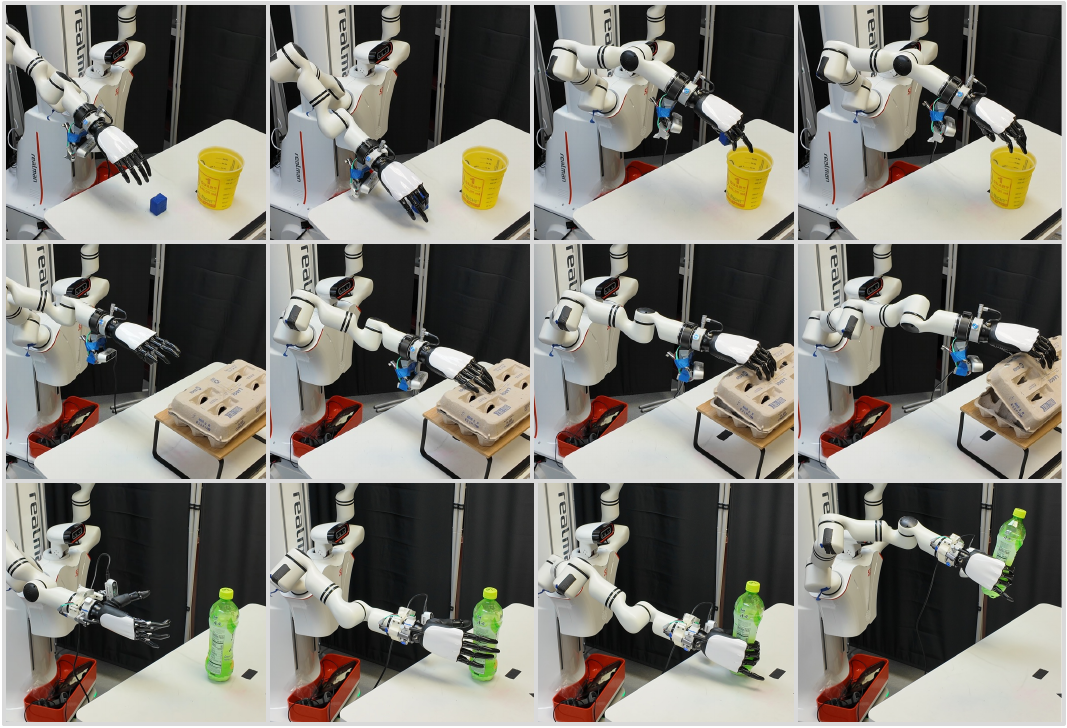}
 \captionsetup{belowskip=-5pt}
 \caption{Policy rollouts for the block pick and place, egg carton, and bottle tasks.}
 \label{fig:policy_rollout}
\end{figure}

\subsection{Demonstration User Studies}

A user study was conducted to compare experience and performance across different demonstration methods. We recruited 14 university students (7 female, 7 male; aged 18--27) with hand sizes ranging from 165~mm to 195~mm. Participants engaged with with 3 demonstration devices: \ACRO, DexUMI, and vision-based teleoperation \cite{rayyan2026teledex}. Teleoperation served as a baseline, as it is the most commonly used approach in prior work. For each device, participants were asked to perform the following tasks:\\
\textbf{Scissors cutting}: Pick up scissors and cut a strip of tape.\\
\textbf{Page flipping}: Use the fingertip to flip a notebook page.\\
\textbf{Cup stacking}: Stack 3 cups facing up.\\
\textbf{Piano playing}: Play 16 notes on a piano using 4 fingers.

For each task, we recorded success rate and completion time as quantitative metrics. All tasks were performed under a 120-second time limit, and completion time was capped at 120 seconds if unfinished. In addition to objective metrics, subjective feedback was collected using Likert-scale questions adapted from NASA-TLX dimensions \cite{hart1988development}. We tested whether \ACRO{} receives higher subjective ratings than DexUMI using a Wilcoxon signed-rank test.

The quantitative results from the user study are shown in Table~\ref{tab:quant_results}. \ACRO was the only device capable of performing the scissors cutting task. DexUMI failed as its added exoskeleton geometry, absent in the original robot hand, prevented the fingers from fitting within the handles. Teleoperation failed at the same task due to a lack of precision, responsiveness, and force feedback. While DexUMI outperforms \ACRO in page flipping and cup stacking in terms of completion time (13.0\% and 25.8\% faster, respectively), \ACRO achieves higher success rate in both tasks. \ACRO outperforms DexUMI significantly in the piano task, with 54.5\% higher success and 16.6\% faster completion time. It is also worth noting that teleoperation performs worse overall compared to both exoskeleton methods.


The subjective feedback results from the user study is shown in Fig.~\ref{fig:subjective_rating}. Participants reported greater finger independence for the exoskeleton design ($p \ll 0.01$), which is consistent with \ACRO{}’s superior performance in the piano task under the quantitative evaluation. \ACRO also received higher ratings in physical comfort ($p=0.0127$) and lower frustration ($p = 0.0219$) compared to DexUMI, which can be attributed to the analytical design considerations to accommodate a wider range of hand sizes, as well as better dexterity from finger independence. Additionally, participants expressed greater willingness to use \ACRO again in future sessions ($p \ll 0.01$), and slightly higher perceived performance rating compared to DexUMI ($p = 0.0544$), supporting our overall hypothesis.


Notably, despite its lowest quantitative performance, teleoperation received the highest ratings in physical comfort and finger independence, attributable to the user's hand remaining unconstrained during operation. Overall, across both quantitative and subjective metrics, \ACRO{} demonstrated the strongest performance among the three devices, while offering improved physical comfort over DexUMI and greater efficiency over teleoperation.

\begin{table}[t]
\centering
\renewcommand{\arraystretch}{1.4} 
\caption{Policy evaluation comparisons across methods and tasks.}
\begin{tabular}{ c | c c c c}
\hline
\multicolumn{1}{c|}{\textbf{Method}} 
& \multicolumn{4}{c}{\textbf{Tasks}} \\
\hline
Finger Condition
& Block & Carton & Bottle &  \\
\hline
No  & \textbf{0.90} & 0.90 & \textbf{0.85}\\
Yes & 0.85 & \textbf{0.95} &  0.80\\
\hline
\end{tabular}
\label{tab:method_tasks}
\end{table}

\subsection{Policy Evaluation}
\label{sec:policy_eval}

We evaluate whether aligned visual and contact geometry embodiment enables effective end-to-end policy learning without visual post-processing, and whether explicit hand-state conditioning remains necessary under this setting.

\paragraph{Experimental Setup}
Policies are trained on demonstrations collected using \ACRO as described in Sec.~\ref{sec:data collection}. The Block task is trained on 200 demonstrations, while Carton and Bottle are trained on 150 demonstrations each. All policies are trained for 300--500 epochs until convergence under identical data splits and augmentation settings. We evaluate three representative manipulation tasks: \\
\textbf{Block:} Grasp a block and place it into a cup, testing precision and fingertip alignment. \\
\textbf{Carton:} Open an egg carton lid using coordinated multi-finger interaction and distributed contact. \\
\textbf{Bottle:} Grasp a bottle and lift it above 50\,mm, highlighting whole-hand grasping with a palm-assisted enclosure. 

For each trained policy, we conduct 20 evaluation trials with randomized object initial poses. Success is defined as complete placement into the cup (Block), opening the lid beyond $30^\circ$ (Carton), and lifting the bottle by at least 50\,mm and holding it stably for 2\,s (Bottle).

\paragraph{Ablation Study}
We ablate the use of explicit hand-state conditioning by comparing policies trained with and without absolute finger pose inputs. All policies share the same diffusion architecture, visual encoder, and training configuration; only the observation inputs differ.

\paragraph{Quantitative Results}
Policy success rates are reported in Table~\ref{tab:method_tasks}, 
with representative policy rollouts shown in Fig.~\ref{fig:policy_rollout}.
For the \textbf{Block} task, success primarily depends on precise fingertip alignment and stable grasp closure during placement. Under wrist-aligned embodiment, finger configuration remains visually observable from RGB input, allowing the policy to infer grasp posture directly from image features.

For the \textbf{Carton} task, which requires coordinated multi-finger interaction and distributed contact, visual cues such as lid deformation and relative hand pose provide sufficient information for closed-loop adjustment, resulting in similar performance with and without explicit finger-state inputs.

For the \textbf{Bottle} task, which emphasizes whole-hand grasping and stable lifting, performance remains similar with and without explicit finger-state conditioning. Because the task primarily relies on gross hand-object alignment and whole-hand grasping, the policy can recover sufficient hand configuration even under major occlusion.

These results suggest that when hardware handles geometric and visual alignment, raw RGB observations provide sufficient state information, making explicit finger-state conditioning redundant.

\paragraph{Comparison to Prior Wearable-Based Pipelines}
We adopt block placement and carton opening tasks to enable comparison with DexUMI. DexUMI reports success rates of 1.00 (Cube) and 0.85 (Carton) under their best configuration using relative actions, image and tactile conditioning, and segmentation with inpainting to mitigate visual embodiment mismatch. 
Under our aligned pseudo-hand embodiment, we achieve 0.90 success on both tasks without segmentation, masking, inpainting, or any tactile feedback as conditioning.

DexUMI's raw image baseline without tactile conditioning or visual post-processing achieves substantially lower success rates, indicating that segmentation and inpainting play a critical role in compensating for embodiment mismatch. In contrast, our hardware-level alignment enables strong performance directly from raw RGB observations while maintaining a substantially simpler end-to-end pipeline.

\paragraph{Discussion}
Overall, these results suggest that hardware-level alignment of geometry, appearance, and viewpoint reduces both human and algorithmic bottlenecks in dexterous learning. By eliminating segmentation and inpainting and reducing reliance on explicit hand-state conditioning, \ACRO enables a streamlined demonstration-to-policy pipeline while retaining competitive task performance.


%% file: sections/Limitations.tex
\section{Limitations}
\label{sec:Discussion}
DexEXO has several limitations worth noting.  First, finger visibility from a top-down viewpoint is partially occluded by the exoskeleton structure, which may affect visual monitoring during certain tasks. Second, the linkage architecture introduces mechanical interference that can limit the range of motion, particularly when interacting with objects on flat surfaces. Third, the pseudo-hand embodiment introduces a slight spatial offset between the operator’s natural hand and the passive hand, which may reduce intuitiveness for first-time users. Finally, although the hardware-level embodiment alignment improves policy transfer, adapting the system to different robot hand form factors requires non-trivial mechanical redesign and integration effort.

In addition, the current system primarily targets demonstration collection for visually guided manipulation with a wrist-mounted camera. Tasks that require significant occlusion handling, multi-view perception, or rich tactile feedback may still benefit from additional sensing modalities such as tactile arrays or depth sensing. Future work will explore integrating multi-modal sensing and improving the mechanical modularity of the system to support rapid adaptation to a broader range of robotic hands and manipulation scenarios.


%% file: sections/S6_Conclusion.tex
\section{Conclusion}
\label{sec:Conclusion}
We presented \textbf{DexEXO}, a wearability-first dexterous exoskeleton designed to enable scalable, cross-operator demonstration collection while preserving structured kinematic correspondence with a target robotic hand. Through analytically modeled finger interfaces and a pose-tolerant thumb mechanism, DexEXO accommodates anthropometric variation without rigid joint alignment or per-user calibration. Experimental validation confirmed consistent IP transmission, structured TM coupling under low-dimensional pose modeling, and substantial residual self-alignment. User studies demonstrated improved comfort and usability relative to prior wearable systems, and policy experiments showed that hardware-level embodiment alignment enables effective end-to-end learning directly from raw wrist-mounted RGB observations. Together, these results suggest that prioritizing wearability and geometric alignment at the hardware level can reduce both human and algorithmic bottlenecks in dexterous robot learning without sacrificing task performance.